\documentclass{article}
\usepackage[T1]{fontenc} 
\usepackage[utf8]{inputenc} 
\usepackage{ismir,amsmath,cite,url}
\usepackage{graphicx}
\usepackage{xcolor}
\usepackage{svg}

\usepackage{array}
\usepackage{booktabs}
\usepackage{multirow}

\usepackage[ruled,linesnumbered]{algorithm2e}

\title{Cue Point Estimation using Object Detection}

\multauthor
{Giulia Arg\"uello \hspace{1cm} Luca A. Lanzend\"orfer \hspace{1cm} Roger Wattenhofer} { 
ETH Zurich\\
{\tt\small \{agiulia, lanzendoerfer, wattenhofer\}@ethz.ch}
}

\def\authorname{G. Arg\"uello, L. A. Lanzend\"orfer, and R. Wattenhofer}

\usepackage[bookmarks=false,pdfauthor={\authorname},pdfsubject={\papersubject},hidelinks]{hyperref}

\begin{document}

\maketitle
\begin{abstract}
Cue points indicate possible temporal boundaries in a transition between two pieces of music in DJ mixing and constitute a crucial element in autonomous DJ systems as well as for live mixing. In this work, we present a novel method for automatic cue point estimation, interpreted as a computer vision object detection task. Our proposed system is based on a pre-trained object detection transformer which we fine-tune on our novel cue point dataset. Our provided dataset contains 21k manually annotated cue points from human experts as well as metronome information for nearly 5k individual tracks, making this dataset 35x larger than the previously available cue point dataset. Unlike previous methods, our approach does not require low-level musical information analysis, while demonstrating increased precision in retrieving cue point positions. Moreover, our proposed method demonstrates high adherence to phrasing, a type of high-level music structure commonly emphasized in electronic dance music. The code, model checkpoints, and dataset are made publicly available.\footnote{\url{https://github.com/ETH-DISCO/cue-detr}}
\end{abstract}
\section{Introduction}\label{sec:introduction}

The skills required by a ``Disc Jockey'' (DJ) are diverse. To record and play live DJ mixes, DJs need to prepare and know their tracks well. An integral part of the track preparation phase is the placement of cue points. Coined by scratch DJs who placed stickers on vinyl records to indicate important sections, the functionality of cue points remains unchanged in the digital setting. A cue point may serve as an annotation for musical highlights, suitable mixing boundaries, or the general track structure which consists of musical phrases. Furthermore, digital cue points allow DJs to quickly loop a track segment or skip back- and forward during a live performance, altering the track structure on the spot. Unfortunately, placing cue points and track preparation is often a cumbersome and time- consuming process. Similarly to other music information retrieval (MIR) tasks, such as onset detection or beat tracking, cue point placement is not straightforward, despite the prominent structural regularity in electronic dance music (EDM)~\cite{UnlockingTheGrooveInEDM}. For instance, the presence of a prelude shifts the track structure, creating irregularity, and similarly, tracks with arbitrary number of additional bars or tempo variations create a significant challenge which needs to be addressed. We therefore ask the question whether cue point estimation can be automated with a learned approach, imitating human cue point placements by training a model on a manually annotated dataset.

This work addresses the placement of cue points, one of the first tasks during the preparation phase of a DJ mix. 
With this goal in mind we present CUE-DETR, a fine-tuned DETR image object detection model trained for cue point estimation on EDM tracks. We show CUE-DETR outperforms previous approaches without requiring detailed and meticulously curated rule sets, which leverage underlying low-level audio information.

Our contributions can be summarized as follows:
\begin{itemize}
    \item We propose CUE-DETR, an object detection model capable of predicting cue points in EDM tracks. Compared to previous methods, our model achieves higher precision and shows significantly closer alignment with manually placed cue points.
    \item We make our EDM-CUE dataset publicly available, which is 35x larger than the previously available cue point dataset~\cite{MDJCUE}. EDM-CUE contains the metadata for 4,710 EDM tracks, which includes tempo, beat, downbeat, and 21k manually placed cue point annotations provided by human experts.
    \item To increase evaluation objectivity, we introduce additional phrase aligned points to evaluate prediction accuracy. Moreover, we open-source the code and model checkpoints to further the research of DJ-related MIR tasks.
\end{itemize}

\section{Related Work}\label{sec:related-work}

Recent years have seen emerging interests in building automated DJ systems where most approaches try to recreate a fully automated DJ pipeline~\cite{HangTheDJ, AutoDJOptimalTempoAdjustment, MusicMixer, AutomatedDNBSystem, HighlightDetectionDJMix, AIDJforEDM, DifferentiableDJMixer}. Such systems aim to create seamless transitions between two tracks, each focusing on a different subset of challenges in the DJ's task pipeline. Cue points are predominantly addressed in the context of finding suitable mix positions in automatic mixing systems~\cite{AutomatedDNBSystem, HighlightDetectionDJMix, HangTheDJ, AutoPlaylistSequencing}. Music structure analysis forms the basis for most cue detection algorithms, as DJ mixes tend to adhere to the underlying high-level track structures~\cite{ComputationalMixAnalysis}. High-novelty regions found through self-similarity~\cite{AudioNoveltySegmentation}, for instance, allow the determination of suitable mix sections based on the high-level music structure~\cite{HeuristicCuePoints, AutoCue, AutomatedDNBSystem}. Furthering the structural knowledge of a track, crowd-sourced scrubbing data from streaming services uncovers additional structural context, as listeners tend to skip forward to the most prominent section of a track~\cite{AutoPlaylistSequencing}. Applying learning-based concepts for the direct search of musical highlights~\cite{HighlightDetectionDJMix} reveals useful information about the musical structure in a similar manner.

Generally, the accuracy of algorithmically chosen cue points varies depending on the granularity and completeness of the rule set implemented in conjunction with the structural analysis~\cite{HeuristicCuePoints}. Adding further rules into the set, for instance, introduces a trade-off between the number of correctly estimated cue point positions and the correctness of each estimated cue point~\cite{AutoCue}. The main focal point of the open-source DJ system Automix~\cite{AutoCue} is a rule-based cue point estimation algorithm, including a validation dataset containing 145 tracks~\cite{MDJCUE}. Automix implements four empirically chosen rules describing possible locations of ``switch points,'' a subset of cue points, on top of structural analysis. Furthermore, the implementation of Automix depends on underlying MIR tasks, such as beat tracking.

DJ mix reverse-engineering~\cite{ReverseEngineeringDJ, DJMixDataExtraction} is a related task to cue point estimation, as it addresses the lack of available and ready-to-use datasets~\cite{ReverseEngineeringDJDataset}. Such ``unmixing'' methods extract latent mixing information from recorded DJ mixes, whose retrieval typically relies on manual annotations, such as mix-in and mix-out points or volume gain curves. The use of pure DJ mix reverse-engineering for cue point estimation is limited as no novel cue points can be retrieved from existing DJ mixes.

In the context of lower-level MIR tasks, convolutional neural networks (CNNs) have been studied, for example, in onset detection~\cite{OnsetDetectionCNN} or beat tracking~\cite{DownbeatTrackingCNN}. Furthermore, CNNs have proven helpful in musical structural analysis and boundary estimation~\cite{BoundaryDetectionCNN}. Using an attention mechanism in conjunction with a CNN can help alleviating the challenges posed by the sequential nature of music. Nevertheless, adding an attention mechanism does not solve the main concern posed by the large amounts of data required for training. Another possible solution is to instead use a large pre-trained model and to then fine-tune the model on task-specific datasets. The Audio Spectrogram Transformer~\cite{AST}, for instance, demonstrates the possibility to transfer a pre-trained ViT model~\cite{ViT} from the image domain to the audio domain. Transformer architectures are often designed to apply the attention mechanism together with a pre-trained CNN backbone, leveraging the feature space previously learned by the CNN~\cite{DETR, SoundEventDetection}.

\section{Methodology}

\subsection{Dataset}

We created EDM-CUE, a dataset containing music metadata from four private collections of professional DJs.
Each of the four DJs uses the library management tool rekordbox\footnote{\url{https://rekordbox.com}} from which we collect the track name, artist name, tempo, beat grid, and cue points for each contained track. Cue points are given by their absolute position in seconds. The beat grid represents a visual metronome, which can be calculated from its stored values: the tempo and grid offset return the beat positions. Applying the time signature in combination with the initial beat number reveals the downbeat. Since we aggregate tracks from four individual collections, all duplicate tracks need to be merged. We summarize the tempo and grid offset to their respective mean values for all duplicate track entries. In order to merge duplicate cue points, we group all cue points based on their distance to neighboring points. Cue points within a distance of a quarter beat of one another form a group. The merged cue point value corresponds to the group center position. All dataset tracks are based on a 4/4 time signature and show constant tempo over time, outlier tracks were excluded during collection. We then pair the information of each track with the track ID found on Deezer\footnote{\url{https://www.deezer.com}} to provide an additional reference.

Our dataset contains 4,710 EDM tracks consisting of around 380 hours of music. The tempo-range lies between 95 and 190 bpm, and track duration ranges from 1 minute 37 seconds to 10 minutes with an average of 4 minutes and 50 seconds. In total, the dataset contains 21,461 cue point annotations with an average count of 4.6 cue points per track. All tracks used to train the model are compressed to 128 kbps MP3 at 44.1 kHz. 

\subsection{Phrasing}\label{sec:phrasing}

\begin{figure}[t]
    \centering
    \includegraphics[width=\columnwidth]{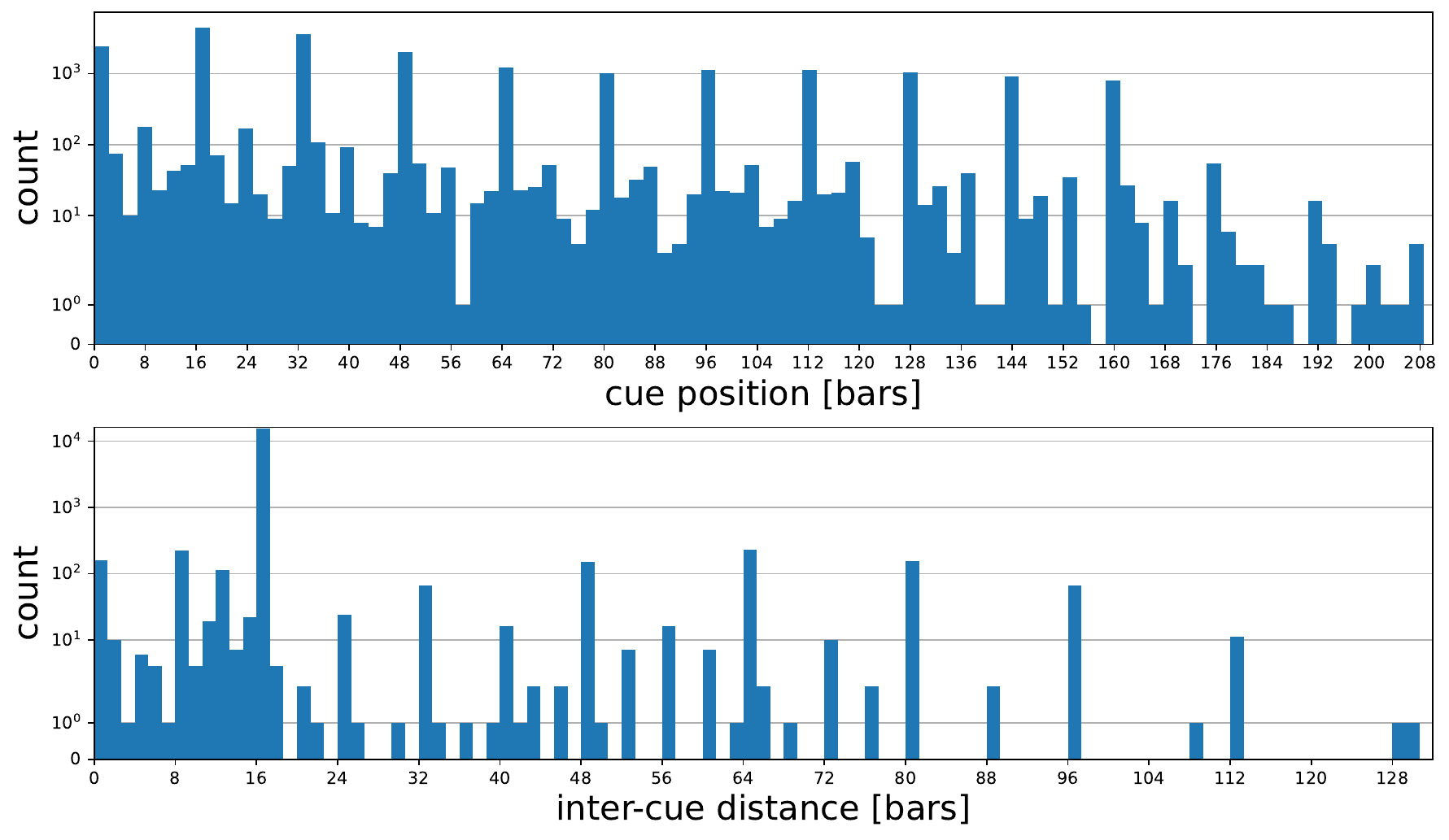}
    \caption{Top: Distribution of cue point positions in EDM-CUE. Bottom: Distribution of distances between two subsequent cue points in EDM-CUE. The inter-cue distances indicate that 16 bars is the most represented phrasing length in our dataset.}
    \label{fig:cue-placements}
\end{figure}

Although cue points frequently align with high-level structural boundaries and tend to strongly coincide with phrase boundaries \cite{ComputationalMixAnalysis}, the placement of cue points is a subjective task with no clear definition; therefore, annotations collected from DJs may not contain all plausible cue points. We first examine the distribution of our training data for cue point positions quantized to bars. Our training cue points exhibit a periodicity with high occurrences of cue points on multiples of 8 and 16 bars, as illustrated in Figure~\ref{fig:cue-placements}.
When also taking the inter-cue spacing between neighboring cue points into account, we observe that a majority of our training tracks adhere to phrase lengths of 16 bars, followed by 8 bars. Due to the strong regularity, we will refer to sections with phrase lengths other than 8 or 16 bars as ``irregular.'' Furthermore, analyzing the cue points in EDM-CUE we find DJs often place cue points at the start of such irregular sections.

\begin{figure}
    \centering
    \includegraphics[trim={2em 1em 1em 0},clip,width=\columnwidth]{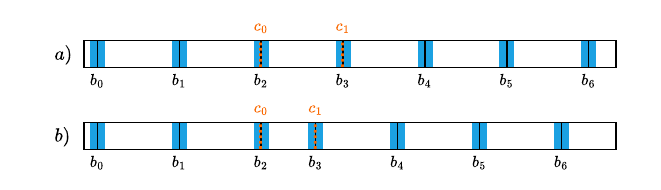}
    \caption{Calculation of phrase boundaries $b_i$ using cue cue points $c_i$. Phrase boundaries, highlighted in blue, serve as additional points to evaluate prediction accuracy. Example a) represents a track with regular phrasing whereas b) shows a track with an irregular phrase between cue points $c_0$ and $c_1$. The computed phrase boundaries $b_i$ include the cue points.}
    \label{fig:phrase-boundary-calc}
\end{figure}

Since regular and clearly defined phrasing is common in EDM~\cite{UnlockingTheGrooveInEDM}, we generalize our collected ground-truth data by estimating phrase boundaries $B$. Phrase boundaries serve as an approximation of the track structure which we use to further validate model accuracy. Using track duration $t$, phrase length $l$, and an ordered, ground-truth cue point set $C$, we find $B$. The non-empty set $C$ must include cue points $c_i$ which mark the start point of irregular phrase boundaries. Traversing the section preceding the first cue point $c_0 = b_0$ in increments of $l$ yields the first entries of $B$. When the iteration reaches a negative value, the remaining track section from $c_0$ is traversed in the opposite direction until $b_i \geq t$. A new boundary $b_i$ is added to $B$ if the iteration step did not skip or reach any $c_i$. Otherwise, the next cue $c_i$ is added to $B$ as $b_i$. The two simplified examples in Figure~\ref{fig:phrase-boundary-calc} show resulting boundaries.

\subsection{Model}
\begin{figure*}[ht!]
    \centering
    \includesvg[width=\textwidth]{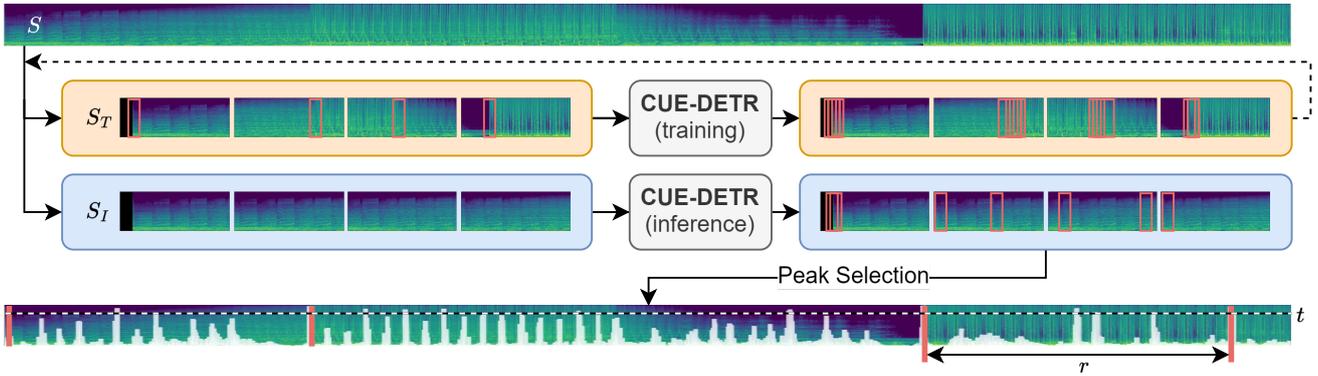}
    \vspace{-0.75cm}
    \caption{Pipeline of the proposed CUE-DETR architecture. During training, an input Mel spectrogram $S$ is segmented into training images $S_T$. Each $S_T$ consists of a spectrogram segment containing a cue point which is represented as a bounding box. Inference images $S_I$ move across $S$ using a sliding window. The predicted bounding boxes are converted to their center $x$-coordinate. The highest scoring positions are selected greedily among all candidates with minimum confidence $t=0.9$. A selected position excludes all other candidates within a radius $r$. The bottom spectrogram shows the predicted positions as peaks based on the confidence value.}
    \label{fig:data-pipeline}
\end{figure*}

Our proposed cue estimation system is based on DETR~\cite{DETR}, a pre-trained object detection transformer. For each track in the dataset, we generate Mel spectrograms using 128 Mel bands at a sampling rate of 22,050 Hz. Our window length measures 2,048 samples, and the hop length is 512 samples.

The input of the model consists of $128 \times 355$ pixel spectrogram segments to fit the expected input image format for DETR while also maximizing the duration of the depicted audio to approximately 11 seconds per image. In the following, we refer to a complete track spectrogram as $S$. The training spectrogram segments $S_T$ and inference spectrogram segments $S_I$ denote the input images of the model. The model returns positional encodings for the predicted bounding boxes alongside the accompanying confidence scores and class labels represented by logits. The data pipeline is illustrated in Figure \ref{fig:data-pipeline}.

\subsection{Preprocessing}
We differentiate between preprocessing for training and inference, as the model is required to process complete spectrograms during inference, whereas for training, the model only requires image segments depicting cue points.
A training image segment $S_T$ is cut from $S$ around a cue point $p$ found in $S$. Using a random integer offset $o \in [0,355)$, image $S_T$ is defined as the segment with left side $p-o$ and right side $p-o+355$. If image $S_T$ partly lies outside of spectrogram $S$, the additional space in $S_T$ is zero-padded. The inclusion of image offset $o$ acts as a simple data augmentation strategy.
For the training annotations, each cue point in an image $S_T$ is encapsulated by a bounding box. The aforementioned box occupies the entire height of $S_T$ and is centered around the cue point. In the event that the box extends beyond the image, it is cropped to align with the image borders. Due to this cropping strategy, all training tracks are split into training and validation sets and are indexed by their respective cue annotations.

To make predictions over the span of a full track, during inference, the complete spectrogram needs to be shown to the model. We employ a sliding window cropping strategy on spectrogram $S$ with an overlap of 0.75 in order to generate inference image segments $S_I$. Similarly to training, the left side of spectrogram $S$ is zero-padded with an arbitrary offset $o \in[89, 266]$ prior to cropping. Applying the zero-padding approaches the uniform distribution of cue point positions seen in the training data, thus increasing the chance to detect cue points at the very start of a spectrogram. As the final step, the resulting image sequence is normalized.

\subsection{Postprocessing}
\begin{figure*}[ht!]
    \centering
    \includegraphics[width=\textwidth,height=4.50cm]{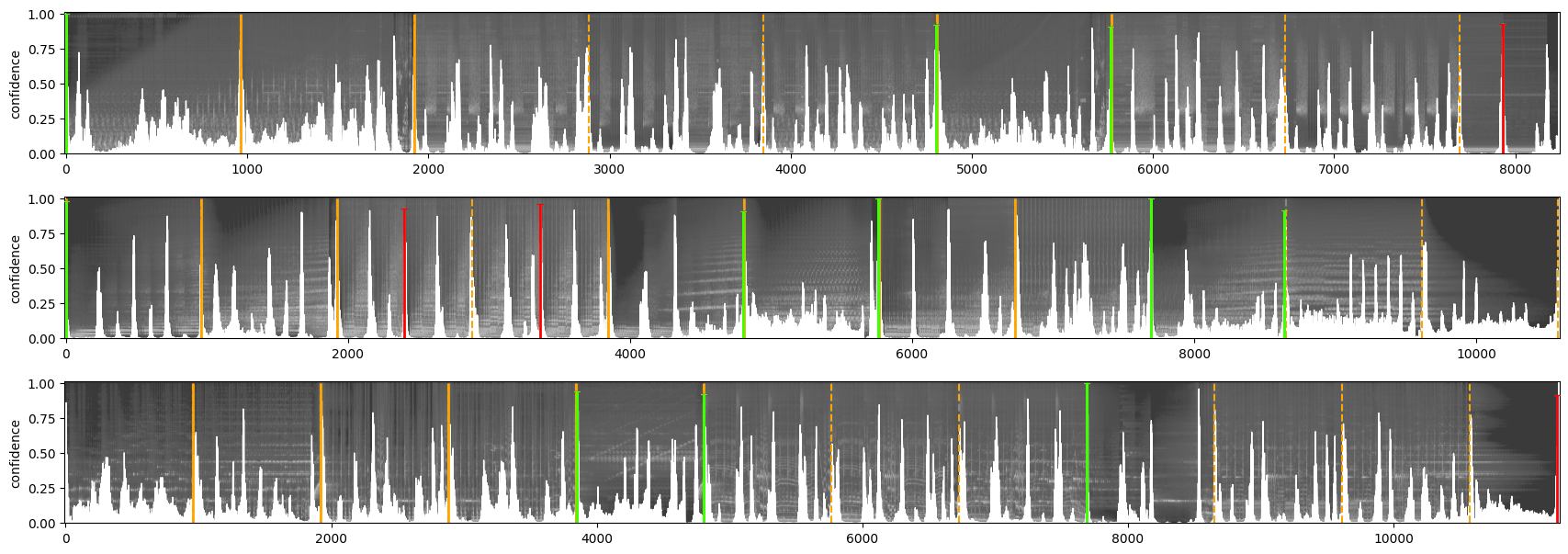}
    \caption{Predicted and ground-truth cue point positions shown over three Mel spectrograms of different random tracks from the evaluation split of EDM-CUE. The confidence score for each position is illustrated as the white curve. Magenta lines indicate correct model predictions, red lines indicate wrong model predictions. For reference, solid orange lines represent ground-truth positions and dashed orange lines illustrate 16-bar phrase boundaries.}
    \label{fig:peaks}
\end{figure*}

We implement additional postprocessing for inference only since additional processing of the basic DETR output is not necessary during training. The model outputs contain the logits and positional encodings mapping to the predicted bounding box coordinates over images $S_I$. Applying a softmax function to the logits yields the class labels and confidence scores for each prediction, cue points are retrieved from the respective positional encodings. The positional box representation is converted to pixel coordinates in corner format to find the center point on the $x$-axis. The resulting point is mapped back to the absolute coordinates of track spectrogram $S$ using the left edge of image segment $S_I$.
Once all conversion results for spectrogram $S$ have been accumulated, the confidence scores are sorted by their associated position, resulting in peaks where the confidence is highest. We implement a peak selection strategy using radius $r$; final cue point candidates are selected in descending order based on their predicted confidence score. Candidates within radius $r$ of a previously selected candidate are ignored. We use a confidence score threshold of 0.9 as the lower bound for selected candidates.
Figure \ref{fig:peaks} shows three example spectrograms with sorted confidence scores. The highest peaks in the curve representation of confidence scores coincide with ground-truth cue points or phrase boundaries, however with noticeable additional high scoring positions. The additional high peaks are predominantly present at 4 bar intervals. As discussed in Section~\ref{sec:phrasing}, we found ground-truth cue points align best with 16-bar phrases. We found that enforcing a minimum spacing $r$ of 16 or 8 bars between consecutively predicted cue points improves the outcome of the final predictions with respect to precision.

\section{Evaluation}

The final evaluation is conducted on 101 tracks which were excluded from the training and validation split. This test set contains 607 ground-truth cue point annotations.

\subsection{Experiment Setup}

We initialize CUE-DETR with pre-trained weights from DETR.\footnote{\url{https://huggingface.co/facebook/detr-resnet-50}} The backbone is initialized with the ResNet-50 weights, and we set the backbone learning rate to $10^{-6}$. For the transformer, we choose a learning rate of $10^{-5}$, and set the weight decay to $10^{-4}$. The bounding box width $w$ is set to 21 pixels and the postprocessing radius $r$ is fixed at 16 and 8 bars, referenced as $r_{16}$ and $r_8$, respectively. We train the model using AdamW~\cite{AdamW} and schedule a learning rate reduction by factor 10 when the validation loss does not improve for 10 epochs. The final model is trained for 50 epochs on one NVIDIA TITAN Xp GPU with a batch size of 192.

While we experimented with training CUE-DETR using randomly initialized transformer weights, we found using pre-trained weights provided significantly better results. Even though the pre-trained transformer weights were trained on COCO 2017~\cite{DETR, DatasetCOCO}, a distinctly different data distribution compared to Mel spectrograms, we corroborate previous findings of visual feature space transfer learning~\cite{AST, Riffusion}.

We compare our model with two other methods, namely ``Mixed In Key 10'' (MIK), a commercial DJ software,\footnote{\url{https://mixedinkey.com}} and Automix~\cite{AutoCue}, an open-source research project. We analyze all tracks directly without manual interference in MIK, as the program simultaneously estimates the beat grid to which it snaps generated cue points. From Automix, we used the cue point generation method directly.

\subsection{Evaluation Metrics}

We investigate the predicted cue points with respect to the manually annotated cue points and phrase alignment separately. In the following, we address the manually annotated cue point ground-truth set by \emph{cues-only} and use the phrase length, measured in bars, to reference phrase alignment. Similarly to Automix, we assess the predictions using a tolerance window around the ground-truth cue points to estimate the hit rate of the predictions. We evaluate the models on two different tolerance windows $T_1$ and $T_{1/2}$ which measure one beat and one half-beat, respectively. On average, one half-beat in our test data measures approximately 172 milliseconds, which is comparable to the standard 150 milliseconds tolerance in beat tracking~\cite{BeatTrackMetric}. The values for precision, recall, the $F_1$-score and Average Precision (AP) scores are retrieved from the hit rate.
Lastly, we measure the cosine similarity between the sets of the predicted and actual cue point positions.

\subsection{Ablations}

As cue points have no clearly defined object boundaries, we further investigate the influence of the spectrogram context around a cue point included in a bounding box. We report the impact of the bounding box width $w$ for the quality of predictions in \tabref{tab:box-ablation} using AP. We report AP for cues-only as $\text{AP}_{C}$ and report AP for phrase alignment as $\text{AP}_{16}$ and $\text{AP}_8$. We trained three models with identical initialization parameters except for $w$ which we set to $w_7=7$, $w_{15}=15$, and $w_{21}=21$ pixels, respectively.

Looking at the results for $T_1$, the box width shows no impact on $\text{AP}_{C}$. The larger peak radius $r_{16}$ increases $\text{AP}_{C}$ for all models. Furthermore, AP increases from $\text{AP}_{C}$ to $\text{AP}_{16}$ for all models, most notably by 0.18 from 0.32 to 0.5 for $w_{21}$ with $r_8$. From $\text{AP}_{16}$ to $\text{AP}_8$ we report an additional increase in AP. Using a larger $w$ improves AP for the phrase alignment cases. The overall best AP score measures 0.6 for $w_{15}$ and $w_{21}$ with radius $r_{16}$. This radius produces identical results for $w_{15}$ and $w_{21}$.
The results for $T_{1/2}$ exhibit similar patterns, with the exception of $w_{21}$ reporting improved AP over $w_{15}$ on all accounts. 
Overall, the model with $w_7$ performs the least favorable, followed by $w_{15}$, which in turn is outperformed by $w_{21}$.

\begin{table}[]
\caption{Ablation of the bounding box width $w$ used during training of CUE-DETR. The Average Precision (AP) scores are reported as $\text{AP}_{C}$ for cues-only ground-truth, $\text{AP}_{16}$ and $\text{AP}_8$ indicate phrase alignment. The best results per scenario are bold and larger values are better.}
\centering
\begin{tabular}{rc|ccc|ccc}
\toprule
\multirow{2}{*}{} & \multirow{2}{*}{}  & \multicolumn{3}{c}{$T_{1}$ (\textit{one beat})} \vline& \multicolumn{3}{c}{$T_{1/2}$ (\textit{half beat})} \\
& $w$ & $\text{AP}_{C}$ & $\text{AP}_{16}$ & $\text{AP}_8$ & $\text{AP}_{C}$ & $\text{AP}_{16}$ & $\text{AP}_8$ \\
\midrule
\multirow{3}{*}{$r_{16}$}& 7 & \textbf{0.41} & 0.51 & 0.52 & 0.34 & 0.37 & 0.37  \\
& 15 & \textbf{0.41} & \textbf{0.57} & \textbf{0.60} & 0.36 & 0.42 & 0.42  \\
& 21 & \textbf{0.41} & \textbf{0.57} & \textbf{0.60} & \textbf{0.38} & \textbf{0.47} & \textbf{0.48} \\
\midrule
\multirow{3}{*}{$r_8$}& 7 & 0.32 & 0.42 & 0.45 & 0.23 & 0.26 & 0.27 \\
 & 15 & 0.32 & 0.49 & 0.53 & 0.25 & 0.33 & 0.34  \\
 & 21 & 0.32 & 0.50 & 0.54 & 0.28 & 0.38 & 0.41  \\
\bottomrule
\end{tabular}
\label{tab:box-ablation}
\end{table}

\subsection{Results}

\begin{table*}[]
    \caption{Comparison of precision, recall, and the $F_1$-score of Automix, Mixed In Key (MIK), and our method. Higher values correspond to better results. The upper rows show the evaluation using tolerance $T_1$ and the lower rows using $T_{1/2}$. From left to right, the results are given for the manually placed cue point only, the computed 16-bar phrasing and the computed 8-bar phrasing. We observe that CUE-DETR outperforms previous methods on precision, recall, and $F_1$-score.}
    \centering
    \begin{tabular}{cl|ccc|ccc|ccc}
         \multicolumn{2}{c}{} & \multicolumn{3}{c}{\textit{cues-only}} & \multicolumn{3}{c}{\textit{16-bars}} & \multicolumn{3}{c}{\textit{8-bars}}\\ 
        \toprule
        & & Precision & Recall & $F_1$ & Precision & Recall & $F_1$ & Precision & Recall & $F_1$ \\
        \midrule
        \multirow{4}{*}{$T_{1}$}
        &Automix & 0.14 & 0.12 & 0.13 & 0.24 & 0.11 & 0.15 & 0.30 & 0.07 & 0.11 \\
        &MIK 10 & 0.20 & 0.25 & 0.22 & 0.21 & 0.13 & 0.16 & 0.25 & 0.08 & 0.12 \\
        &CUE-DETR ($r_{16}$) & \textbf{0.38} & 0.35 & 0.36 & \textbf{0.62} & 0.27 & 0.38 & \textbf{0.69} & 0.16 & 0.26 \\
        &CUE-DETR ($r_8$) & 0.32 & \textbf{0.49} & \textbf{0.39} & 0.53 & \textbf{0.41} & \textbf{0.46} & 0.63 & \textbf{0.26} & \textbf{0.36} \\
        
        \midrule
        \multirow{4}{*}{$T_{1/2}$}
        &Automix & 0.11 & 0.10 & 0.10 & 0.20 & 0.09 & 0.13 & 0.24 & 0.06 & 0.10 \\
        &MIK 10 & 0.14 & 0.19 & 0.16 & 0.15 & 0.09 & 0.12 & 0.18 & 0.06 & 0.09 \\
        &CUE-DETR ($r_{16}$) & \textbf{0.27} & 0.25 & 0.26 & \textbf{0.43} & 0.19 & 0.27 & \textbf{0.48} & 0.11 & 0.18 \\
        &CUE-DETR ($r_8$) & 0.22 & \textbf{0.34} & \textbf{0.27} & 0.37 & \textbf{0.28} & \textbf{0.32} & 0.43 & \textbf{0.17} & \textbf{0.25} \\
        \bottomrule
        \end{tabular}
    \label{tab:comparison-prf}
\end{table*}

\begin{figure}[t]
    \centering
    \includegraphics[width=\columnwidth]{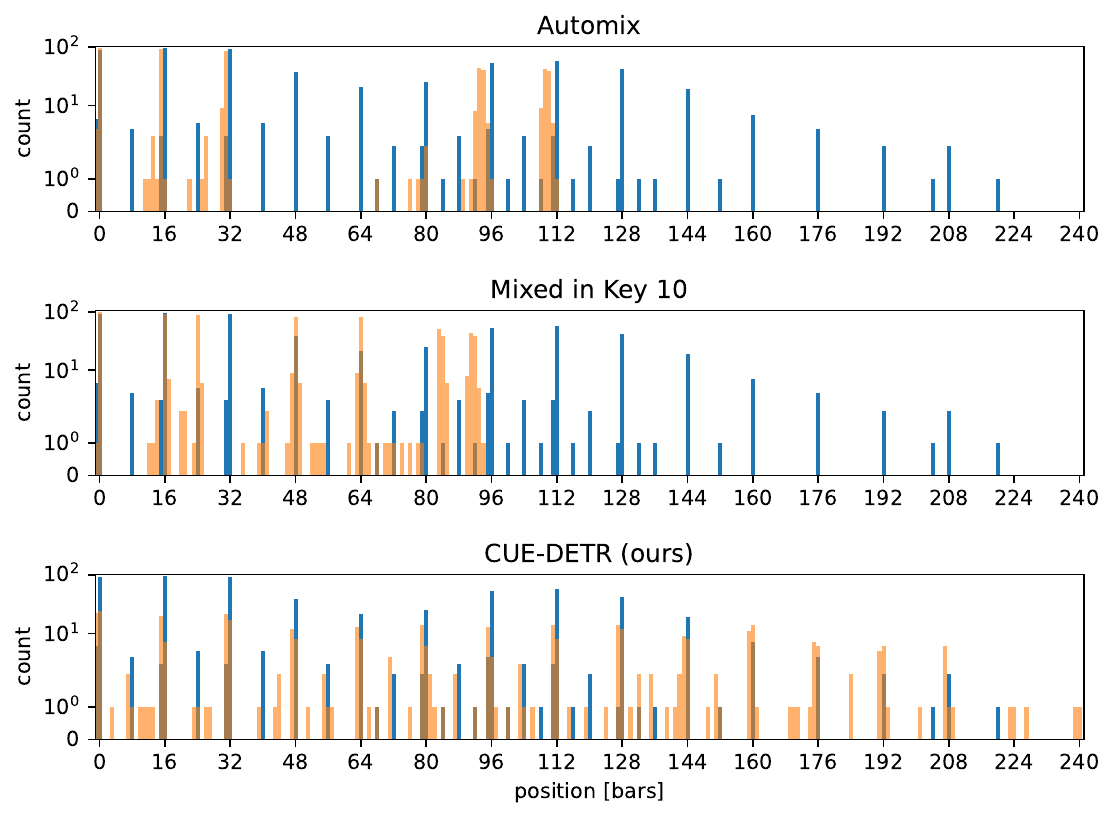}
    \caption{Distribution of ground-truth cue point positions in blue and predicted cue point positions in orange quantized to bars. The cosine similarity between the predicted cue point positions and ground-truth is 0.425 (Automix), 0.371 (MIK), and 0.851 (CUE-DETR).}
    \label{fig:cue-position-bars}
\end{figure}

The evaluation of the mean precision, recall, and the $F_1$-score is summarized in \tabref{tab:comparison-prf}.
For all methods, the precision increases from the cues-only to the 16-bars and 8-bars ground-truth sets. Our $r_{16}$-model achieves the highest precision in all cases. The precision increases most notably for tolerance $T_1$ from the cues-only to phrase alignment ground-truth sets. More precisely, our $r_8$-model shows an increase in precision by 0.31 from cues-only to 8-bar phrasing. The change from 16 to 8-bars is not as prevalent. Automix shows an improvement in precision from 0.14 to 0.24 and 0.3 over the three ground-truth sets. MIK shows little improvement over the different scenarios and produces more stable precision values. Using the tighter tolerance $T_{1/2}$, all precision values fall in proportion to each other. 
For recall, the difference of values between the two tolerances is similar to what is observed for precision. With the added phrasing boundaries, all methods show a reduction in recall, opposite to precision. The most significant drop in recall is observed from 16 to 8-bars. Our $r_8$-model reports the highest recall on all accounts. 
The changes in the $F_1$-score are less pronounced for all methods as the values remain nearly stable for cues-only and 16-bar phrase alignment. The best reported $F_1$-score is associated with our $r_8$-model over 16-bar phrasing at 0.46.

For further insight, we look at the distribution of the predicted results in Figure~\ref{fig:cue-position-bars}. Automix favors cue positions around the first three phrases with high alignment to ground-truth. The second cluster is predicted at the start of phrases 6 to 8 with an increased tendency for early predictions. MIK on the other hand exhibits more evenly distributed cue placements over the first 7 phrases. However, an increased number of predictions lie in between phrases where no ground-truth points lie. We observe that both Automix and MIK tend not to predict possible cue points in the second half of tracks.
CUE-DETR predicts cue points with the highest adherence to ground-truth. Despite a few additional predictions similar to MIK, positions with the highest accumulation of cue points are covered by our predictions in a similar pattern. The cosine similarity of our quantized predictions reports the highest score of 0.851. In comparison, Automix scores 0.425 whereas MIK reaches 0.371.

\subsection{Discussion}

CUE-DETR shows strong adherence to ground-truth compared to other methods. Our method suggests good phrase alignment based on the distribution of our predicted cue point positions, as well as the increase in precision from cues-only to 16 bar phrases. A slight increase in precision is expected for all methods, however, a significant increase is only associated with strong phrase alignment due to the decrease in false positive predictions. The higher number of possible ground-truth positions decreases recall in return. If our method successfully detects irregular sections, the phrasing algorithm from Section \ref{sec:phrasing} can be applied in postprocessing, which could further increase the precision while keeping the recall score high.

Despite using a metronome-agnostic approach, for which we fixed the distances $r$ to the length of a phrase in terms of the dataset median tempo, the chosen values for $r$ yield results with higher precision compared to the other methods. We assume the relatively homogeneous nature of our dataset minimized the impact of different tempos in the test data. For more diverse styles of music, including the tempo and beat grid information, similar to MIK, might be beneficial.
On the other hand, it might be possible to train a model on beat and cue detection simultaneously. The beat detection could then be used during postprocessing to identify the tempo, making the need for additional ground-truth beat grid or tempo information redundant.

One key limitation remains in the availability of training data, despite building our own dataset. Since we only had access to data with high similarity in style, we would like to investigate the performance of our method over a broader domain of electronic music in the future. Furthermore, our dataset annotations were provided by DJs who specialize in club DJing. Therefore, annotations from other types of DJs, such as scratch DJs or mobile DJs, would likely result in a largely different cue point distribution. We believe one main difference would lie in more cue points distributed around vocals or pickups instead of the first downbeat of phrases.

\section{Conclusion}

In this work we introduced CUE-DETR, an object detection model fine-tuned on Mel spectrograms capable of estimating cue points in EDM tracks.
Candidate cue points produced by CUE-DETR demonstrate high adherence to the underlying music structure and exhibit a higher resemblance to manually placed cue points compared to previous approaches.
Furthermore, we created EDM-CUE, a dataset containing 21k manually annotated cue points from four professional DJs. EDM-CUE also contains tempo, beat, and downbeat annotations for almost 5k EDM tracks.
Our implementation includes a postprocessing step to filter the model predictions for the best positions, including a conversion of the results to timestamps. For the evaluation, we presented a complementary phrasing-based evaluation method, which is useful to assess cue point predictions in a more objective manner.

Furthermore, we demonstrated that CUE-DETR is capable of detecting large structural boundaries in music, despite only seeing small excerpts of the entire track. Our findings further acknowledge the potential of transformer-based architectures for the detection of time-based events in music.

\bibliography{references}

\end{document}